\documentclass{article}
\usepackage{spconf,amsmath,graphicx}
\usepackage{cite}
\usepackage{color,soul}
\usepackage{amssymb}
\usepackage{subfig}
\usepackage[colorlinks]{hyperref}
\hypersetup{
  colorlinks=true,
  citecolor=blue,
  linkcolor=red,
  urlcolor=blue}
\usepackage{setspace}
\usepackage{color,soul}

\usepackage{etoolbox}
\patchcmd{\thebibliography}
  {\settowidth}
  {\setlength{\itemsep}{0pt plus -1ex}\settowidth}
  {}{}

\usepackage{booktabs}
\usepackage{colortbl}       
\usepackage{xcolor}

\newcommand{\midsepremove}{\aboverulesep = 0.ex \belowrulesep = 0.ex}
\midsepremove
\newcommand{\midsepdefault}{\aboverulesep = 0.2ex \belowrulesep = 0.2ex}
\midsepdefault

\title{Regression or Classification? New Methods to Evaluate No-Reference Picture and Video Quality Models}
\name{\begin{tabular}{c}Zhengzhong~Tu$^{1\star}$\thanks{$^\star$Equal contribution},~Chia-Ju~Chen$^{1\star}$,~Li-Heng~Chen$^{1}$,~Yilin~Wang$^2$,~Neil~Birkbeck$^2$,\\
~Balu~Adsumilli$^2$,~and~Alan~C.~Bovik$^1$\end{tabular}}
\address{$^1$~The~University~of~Texas~at~Austin,~$^2$~Google~Inc.}
\begin{document}
%
\maketitle
\begin{abstract}
Video and image quality assessment has long been projected as a regression problem, which requires predicting a continuous quality score given an input stimulus. However, recent efforts have shown that accurate quality score regression on real-world user-generated content (UGC) is a very challenging task. To make the problem more tractable, we propose two new methods - binary, and ordinal classification - as alternatives to evaluate and compare no-reference quality models at coarser levels. Moreover, the proposed new tasks convey more practical meaning on perceptually optimized UGC transcoding, or for preprocessing on media processing platforms. We conduct a comprehensive benchmark experiment of popular no-reference quality models on recent in-the-wild picture and video quality datasets, providing reliable baselines for both evaluation methods to support further studies. We hope this work promotes coarse-grained perceptual modeling and its applications to efficient UGC processing.

\end{abstract}
\section{Introduction}
\label{sec:intro}

The success of social media as an industry, coupled with the expansion of video traffic on the Internet in recent years, is driving a continuous focus on video/image processing and streaming. Video compression makes streaming possible, while video quality models such as PSNR, SSIM \cite{wang2004image}, and VMAF \cite{li2018vmaf}, which measure perceptual differences between original and compressed videos, serve to calibrate trade-offs between rate and quality in compression. These usually operate under the assumption that original videos have pristine quality. However, this presumption is not true for media sharing platforms like YouTube and Facebook, since the majority of uploaded videos are user-generated content (UGC), which often already suffers from unpredictable quality degradations, commonly incurred during capture. In this case, the original quality of UGC, which can only be measured by no-reference quality models, must also be included as an important factor when optimizing UGC compression or transcoding.

Many blind video quality assessment (BVQA) models have been proposed to solve this `UGC-VQA' problem \cite{tu2020ugc, mittal2012no, ghadiyaram2017perceptual, saad2014blind, korhonen2019two, li2019quality, ebenezer2020no}, among which a simple but effective strategy is to compute frame-level quality scores, e.g., as generated by blind image quality assessment (BIQA) models \cite{mittal2012no,xue2014blind,ye2012unsupervised, ghadiyaram2017perceptual, tu2020bband}, followed by some form of temporal quality pooling \cite{seshadrinathan2011temporal, tu2020comparative, li2019quality, chen2020perceptual}. Other recent methods leverage end-to-end training of convolutional neural networks to predict quality scores \cite{ma2017end, li2019quality, kang2014convolutional, Bosse2018}. Either way, the BVQA/BIQA problem has nearly always been cast as a regression problem, where a continuous quality score is predicted from a given visual signal. The success of these models is evaluated by comparing their quality predictions to subjective mean opinion scores (MOSs) \cite{sheikh2006statistical}, which are usually collected by conducting large-scale human studies.


Here we study the no-reference quality assessment of UGC (UGC-QA) in a new light, and propose alternative approaches to the well-established regression approach. The UGC-QA problem is similar to the image aesthetics assessment (IAA) problem \cite{deng2017image, Murray2012, zeng2019unified}, as they both seek to predict subjectivity and then share various intertwined factors. UGC-QA focuses more on technical quality such as distortion, rather than what makes a picture aesthetically appealing. Inspired by the formulation of IAA problems, we quantize the original subjective labels (MOSs) with different degrees of granularity, onto 1) binary labels for binary \texttt{high} vs. \texttt{low} quality classification, and 2) ternary labels for finer-grained quality categorization. These evaluation methods relax the use of continuous labels in the original regression task.

There are at least three good reasons to take this approach. First, recent work has shown that accurate quality score regression is a very challenging task \cite{tu2020ugc, ying2020patches}, and even the state-of-the-art models suffer considerable prediction uncertainty \cite{tu2020ugc}. Like the IAA problem \cite{Murray2012}, relaxing regression to (binary) classification could hence make this problem more tractable. Second, inspired by just-noticeable-difference (JND) \cite{lin2015experimental} approaches to reference-based video quality, we suggest that for blind visual quality prediction, similar JND-like approaches may be taken to exploit the visual discriminative power and limits of subjects' quality perception, as exemplified by the popular five-level absolute category rating (ACR) scale \cite{huynh2010study}: \{\texttt{Bad}, \texttt{Poor}, \texttt{Fair}, \texttt{Good}, \texttt{Excellent}\}. Discretizing continuous scores onto quality categories follows this method. Finally, such an approach may be more useful on UGC media transcoding platforms like YouTube and Facebook, i.e., classifying quality scores onto categories, since only discrete quality-guided decisions can be deployed when (pre-)processing UGC content. One example that helps explain this is the quality-guided transcoding (QGT) framework proposed in \cite{Wang2020}, which involves encoding videos uploaded to YouTube with parameters optimized based on its input quality category: \{\texttt{low}, \texttt{medium}, \texttt{high}\}. To this end, it is also of interest to determine the capability of a model to predict quality at coarse levels by evaluating the accuracy of classification tasks instead of regression. 

\section{Problem and Task Formulation}
\label{sec:form}

We first revisit the formulation of the classic UGC-QA regression problem, and then discuss the relaxed classification tasks we propose. Some evaluation metrics are also presented for each individual task.

\begin{figure}[!t]
\centering
\footnotesize
\def\imghei{0.4\textwidth}
\begin{tabular}{c}
\includegraphics[width=\imghei]{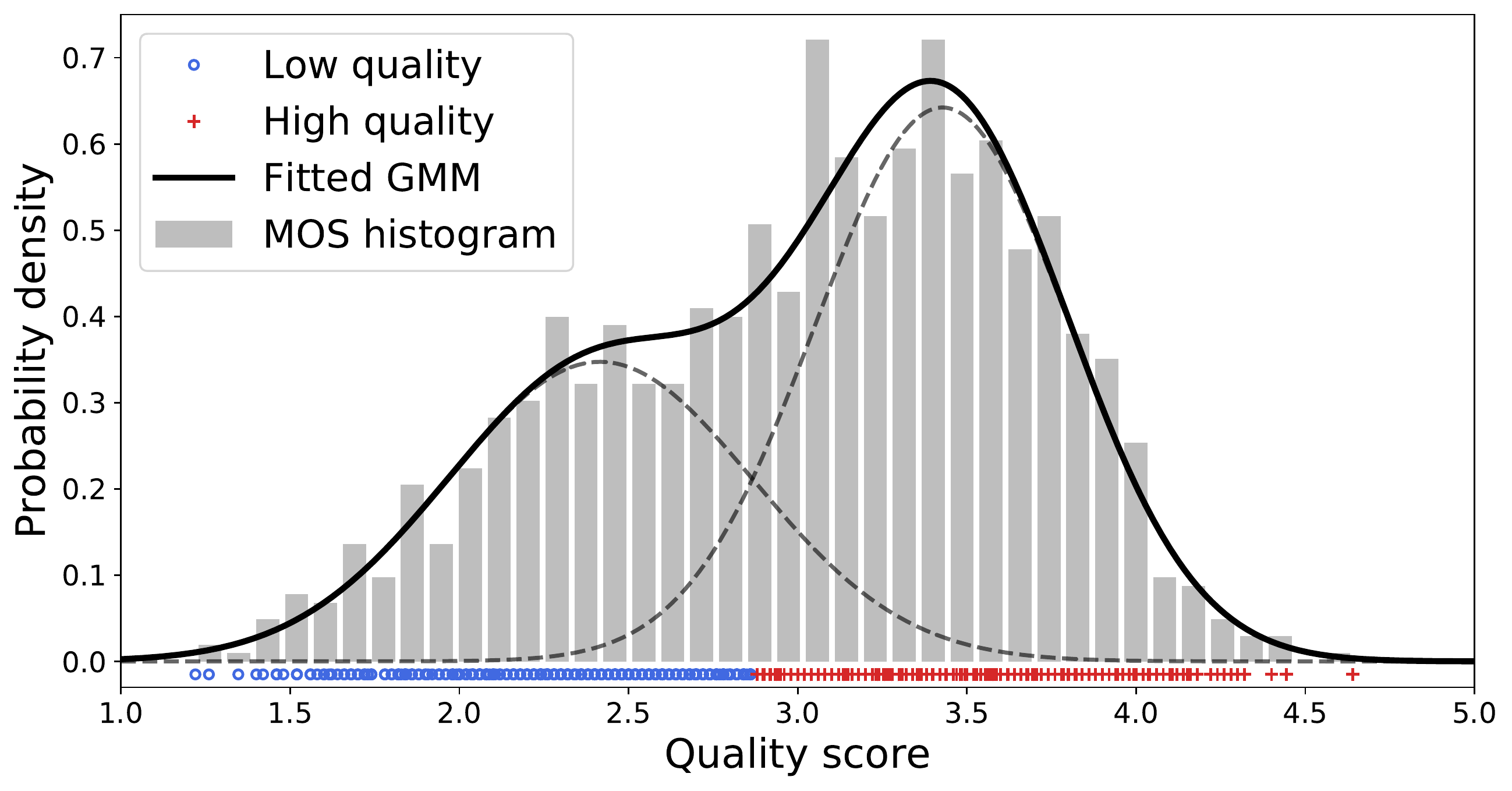}  
\end{tabular}
\caption{MOS distribution of KoNViD-1k \cite{hosu2017konstanz} follows a multimodal distribution. Here, an example of a mixture model of two Gaussians is fit to cluster the samples into \texttt{high} (red) or \texttt{low} (blue) quality categories for binary classification.}
\label{fig:dist}
\end{figure}

\subsection{Task A: Regression}
\label{task1}

Consider a set of training samples $\mathbb{D}=\{(\mathbf{x}_i,y_i)\}_{i=1}^m$, where $\mathbf{x}_i\in\mathcal{X}$ and  $y_i\in\mathcal{Y}\in[a,b]\in\mathbb{R}$, i.e., $\mathbf{x}$ is either in the picture or video space and $y$ is a continuous quality score in an interval. The supervised regression task, which is the standard treatment for UGC-QA problem, is to find a hypothesis or model $h:\mathcal{X}\rightarrow\mathcal{Y}$ that best approximates the true relationship between variables and targets. Classical feature-based models \cite{tu2020ugc, mittal2012no, ghadiyaram2017perceptual, saad2014blind, korhonen2019two, ebenezer2020no} basically involve two steps - first manually design a feature extractor $\Phi$ that maps the raw pixel space to a much smaller yet informative feature space: $\Phi: \mathcal{X}\rightarrow \mathcal{Z}$ ($\dim(\mathcal{Z})\ll\dim(\mathcal{X})$), then learn a shallow regressor, e.g., support vector machines (SVM) \cite{chang2011libsvm}, or random forests (RF) \cite{pei2015image}, in the transformed domain $\{(\mathbf{z}_i,y_i)\}_{i=1}^m$. Recent end-to-end solutions \cite{kang2014convolutional, Bosse2018, li2019quality, chen2019proxiqa} jointly learn, from the raw pixel domain, feature representation and regression layers within a convolutional neural networks by optimizing $\ell_p\ (p=1,2)$ losses between predicted scores and the ground truth.

The standard performance metrics for UGC-QA regression are the Spearman rank-order correlation coefficient (SRCC) calculated between the ground truth MOSs and the predicted scores to measure the prediction monotonicity, and the Pearson linear correlation coefficient (PLCC) to measure the degree of linear correlation against MOS, sometimes in company with root mean squared error (RMSE). 

Regression inherently imposes that the output space is a metric space, where it penalizes the prediction error uniformly over the entire output range. However, we suggest that there may exist quality thresholds to break up the continuous quality range into semantic quality categories, which is conceptually similar to the JND approaches \cite{lin2015experimental} that measures the quality loss of compression. That being said, it is somewhat reasonable to deliberately discretize the continuous scores into $N$ discrete bins, based on the assumption that samples lying within the same bin have very similar perceptual qualities. The smaller $N$, the easier the resulting task, at the cost of more relaxation of labels.

It has been observed that the empirical distribution of MOSs on a UGC-QA dataset usually follows a unimodal \cite{ghadiyaram2015massive, hosu2020koniq, yu2019predicting} or multimodal \cite{yu2020predicting, chen2020perceptual, hosu2017konstanz, sinno2018large, wang2019youtube} distribution, and the authors of \cite{yu2019predicting} have also shown that the score distribution at different distortion levels is often roughly normally distributed. Therefore, we assume that the distributions of quality scores can be represented by a Gaussian mixture model (GMM) with $N$ components as:
\setlength{\belowdisplayskip}{2pt} \setlength{\belowdisplayshortskip}{2pt}
\setlength{\abovedisplayskip}{2pt} \setlength{\abovedisplayshortskip}{2pt}
\begin{equation}
\label{GMM}
p(y)=\sum_{n=1}^N\pi_n \mathcal{N}(y|\mu_n,\sigma_n^2),
\end{equation}
where each Gaussian component presents one semantic class out of $N$ categories. Expectation minimization (EM) is employed as a maximum likelihood estimator to fit the density of the GMM, based on which scores are assigned to a single cluster using the predicted posterior probabilities. Fig. \ref{fig:dist} shows the MOS distribution of the KoNViD-1k \cite{hosu2017konstanz} dataset, where the MOS histogram may be modelled as following a multimodal density function. Applying a mixture model with two Gaussians, which are fit to the histogram, allows clustering the data into \texttt{low} and \texttt{high} quality classes. 

\subsection{Task B: Binary Classification}
\label{task2}

When the number of classes $N$ is chosen as $2$, the original regression problem reduces to a binary classification task, i.e., to predict whether an input UGC belongs to the \texttt{High} or \texttt{Low} quality category. The binarizing threshold $T$ is automatically determined by the GMM clustering described above. 
This binary categorization problem is particularly interesting since it caters to applications involving optimizing transcoding configurations for low and high input quality separately, such as the QGT framework \cite{Wang2020}. An alternative way of achieving binary predictions is to fit a regressor to the MOS labels, then apply a threshold to obtain binary predictions. Our preliminary experiments on CLIVE \cite{ghadiyaram2015massive}, however, show that the regression-thresholding method achieves worse results than the proposed binary classification training scheme.

Typical evaluation metrics for binary classification include the overall accuracy metric: $\mathrm{Acc.}=\frac{TP+TN}{P+N}$, where $TP$, $TN$, $P$, $N$ denote true positive, true negative, total positive, and total negative, respectively. This metric alone, however, could be biased towards a dominant class. To complement this metric when benchmarking on imbalanced testing sets, the balanced accuracy score can be used: $\mathrm{Balanced\ Acc.}=\frac{1}{2}\left(\frac{TP}{TP+FN}+\frac{TN}{TN+FP}\right)$,
where $FN$ and $FP$ are false negative and false positive.

\subsection{Task C: Ordinal Classification}
\label{task3}

Binary quality categorization is the coarsest classification task, which may not accommodate applications where more than two decisions are preferred. Thus, we also consider $N>2$ to allow for finer-grained single-label multi-class classification. To quantize the quality interval, e.g., $[1,5]$, onto a set of representative discrete labels, we also adopt the GMM clustering with $N$ Gaussians to fit the MOS histogram.

\newcommand{\ra}[1]{\renewcommand{\arraystretch}{#1}}

\begin{table}[!t]
\footnotesize
\ra{1.1}
\setlength{\tabcolsep}{1.6pt}
\caption{Summarization of the benchmarked UGC-IQA (top three rows) and UGC-VQA (bottom three rows) datasets.}
\label{table:1}
\centering
\begin{tabular}{ lllllll }
\toprule
 Database & \# Cont. & Label & Range & Thr. (B) & Thrs. (Task C) \\ \midrule
 CLIVE'16 \cite{ghadiyaram2015massive}  & 1,162  &  MOS+$\sigma$ & [0,100] & \{49.426\} & \{36.929,61.478\} \\
 \rowcolor[gray]{.92}
 KonIQ-10k'18 \cite{hosu2020koniq} & 10,073 &  MOS+$\sigma$ & [1,5] & \{2.9631\} & \{2.5902,3.2688\} \\
 SPAQ'20 \cite{fang2020perceptual}  & 11,125 &  MOS & [0,100] & \{51.475\} & \{38.980,59.475\} \\ \midrule
KoNViD-1k'17 \cite{hosu2017konstanz}  & 1,200  & MOS+$\sigma$ & [1,5] & \{2.8549\} & \{2.5988,3.2900\}\\
\rowcolor[gray]{.92}
 LIVE-VQC'18 \cite{sinno2018large}  & 585    & MOS & [0,100] &  \{57.948\} & \{48.211,67.265\}  \\
  YT-UGC'20 \cite{wang2019youtube}  & 1,380  & MOS+$\sigma$ & [1,5] & \{3.4765\} & \{3.0490,3.9430\} \\
 \bottomrule
\end{tabular}
\end{table}

These class labels are imbued with order information, e.g., the five-point classes include a natural label ordering: $\mathtt{Bad}\prec\mathtt{Poor}\prec\mathtt{Fair}\prec\mathtt{Good}\prec\mathtt{Excellent}$, where $\prec$ is an order relation. In other words, misclassification costs are not the same for different errors, e.g., misclassifying $\mathtt{Bad}$ as $\mathtt{Excellent}$ should be more penalized than misclassifying as $\mathtt{Fair}$. Therefore, we formulate the task as an ordinal classification (or original regression) problem, in which a sample must be classified into exactly one of the five ordered classes.

Several measures can be considered when evaluating ordinal classification models \cite{gutierrez2015ordinal}, among which we choose two common methods: mean zero-one error (MZE) and the mean absolute error (MAE). MZE is the global error rate of the classifier without considering the order: $
\mathrm{MZE}=\frac{1}{N}\sum_{i=1}^N[\![ \hat{y_i}\neq y_i^\mathrm{OC}]\!]=1-\mathrm{Acc.}$, 
where $y_i^\mathrm{OC}$ is the true label, $\hat{y_i}$ is the predicted label and $\mathrm{Acc}$ is the accuracy of the classifier. The $\mathrm{MAE}$ is the average absolute deviation between the predicted rank $\mathcal{O}(\hat{y_i})$ and the true rank $\mathcal{O}(y_i^\mathrm{OC})$: $
\mathrm{MAE}=\frac{1}{N}\sum_{i=1}^N |\mathcal{O}(\hat{y_i})-\mathcal{O}(y_i^\mathrm{OC})|$.



\begin{table*}[!t]
\footnotesize
\ra{1.1}
\setlength{\tabcolsep}{1.4pt}
\caption{Performance comparison of BIQA models on three UGC-IQA datasets. The underlined and boldfaced entries indicate the best and top three performers on each database, for each performance metric of each task, respectively.}
\label{table:iqa}
\centering
\begin{tabular}{ l||cccccc|cccccc|cccccc}
\toprule
Database & \multicolumn{6}{c|}{CLIVE \cite{ghadiyaram2015massive}} &  \multicolumn{6}{c|}{KonIQ-10k \cite{hosu2017konstanz}}
&  \multicolumn{6}{c}{SPAQ \cite{fang2020perceptual}} 
\\ \cmidrule(lr){2-7}\cmidrule(
lr){8-13}
\cmidrule(lr){14-19}
BIQA task & \multicolumn{2}{c}{Regression} &  \multicolumn{2}{c}{Binary Class.} &  \multicolumn{2}{c|}{Ordinal Class.} & \multicolumn{2}{c}{Regression} &  \multicolumn{2}{c}{Binary Class.} &  \multicolumn{2}{c|}{Ordinal Class.} 
& \multicolumn{2}{c}{Regression} &  \multicolumn{2}{c}{Binary Class.} &  \multicolumn{2}{c}{Ordinal Class.} 
\\ \cmidrule(lr){2-3}\cmidrule(lr){4-5}\cmidrule(lr){6-7}\cmidrule(lr){8-9}\cmidrule(lr){10-11}\cmidrule(lr){12-13}
\cmidrule(lr){14-15}\cmidrule(lr){16-17}\cmidrule(lr){18-19}
Model & SRCC$\uparrow$ & PLCC$\uparrow$ &   Acc.$\uparrow$ &  B.Acc.$\uparrow$ &  MZE$\downarrow$ & MAE$\downarrow$ & SRCC$\uparrow$ & PLCC$\uparrow$ &   Acc.$\uparrow$ &  B.Acc.$\uparrow$ &  MZE$\downarrow$ & MAE$\downarrow$ 
& SRCC$\uparrow$ & PLCC$\uparrow$ &   Acc.$\uparrow$ &  B.Acc.$\uparrow$ &  MZE$\downarrow$ & MAE$\downarrow$ 
\\\midrule
BRISQUE 
& .592 & .620 & 75.7 & 69.9 & .416 & .482
& .709 & .715 & 82.0 & 78.0 & \textbf{.346} & \textbf{.372}
& .807 & .814 & 85.9 & 85.9 & .288 & .325
 \\
GM-LOG 
& .599 & .618 & 75.8 & 69.6 & .416 & .480
& .714 & .721 & 82.2 & 77.3 & .351 & .373
& .820 & .825 & 86.1 & 86.1 & \textbf{.286} & \textbf{.322}
\\

HIGRADE
& .622 & .638 & 77.1 & 72.0 & .435 & .505
& \textbf{.781} & \textbf{.799} & \textbf{85.5} & \textbf{81.0} & \textbf{.317} & \textbf{.336}
& \textbf{.855} & \textbf{.860} & \textbf{87.8} & \textbf{87.9} & \textbf{.271} & \textbf{.304}
\\
FRIQUEE
& \textbf{.677} & \textbf{.704} & \textbf{\underline{80.4}} & \textbf{\underline{74.3}} & \textbf{.399} & \textbf{.463}
& \textbf{\underline{.821}} & \textbf{\underline{.839}} & \textbf{\underline{86.9}} & \textbf{\underline{83.4}} & \textbf{\underline{.295}} & \textbf{\underline{.321}} 
& \textbf{.886} & \textbf{.891} & \textbf{\underline{89.6}} & \textbf{\underline{89.7}} & \textbf{\underline{.234}} & \textbf{\underline{.252}}
 \\

CORNIA 
& .644 & .683 & 77.3 & 70.7 & \textbf{.412} & \textbf{.480}
& .730 & .760 & 84.4 & 80.1 & .362 & .389  
& .796 & .804 & 85.4 & 85.4 & .317 & .357
\\
HOSA
& \textbf{.657} & \textbf{.689} & \textbf{80.1} & \textbf{73.7} & \textbf{\underline{.395}} & \textbf{\underline{.436}}
& .673 & .708 & \textbf{85.1} & \textbf{81.8} & .358 & .384 
& .840 & .847 & 84.9 & 84.9 & .314 & .343
\\

VGG-19 
& .587 & .640 & 77.2 & 69.3 & .465 & .588
& .685 & .715 & 80.7 & 74.2 & .351 & .387 
& .807 & .816 & 83.5 & 83.6 & .299 & .342
\\
ResNet-50 
& \textbf{\underline{.701}} & \textbf{\underline{.742}} & \textbf{79.4} & \textbf{72.7} & .457 & .572
& \textbf{.805} & \textbf{.838} & \textbf{85.5} & 80.6 & .369 & .437 
& \textbf{\underline{.889}} & \textbf{\underline{.894}} & \textbf{88.9} & \textbf{89.0} & .295 & .350

\\
 \bottomrule
\end{tabular}
\end{table*}

\begin{table*}[!t]
\footnotesize
\ra{1.1}
\setlength{\tabcolsep}{1.35pt}
\caption{Performance comparison of BVQA models on the three UGC-VQA datasets. The underlined and boldfaced entries indicate the best and top three performers on each database, for each performance metric of each task, respectively.}
\label{table:vqa}
\centering
\begin{tabular}{l||cccccc|cccccc|cccccc}
\toprule
Database & \multicolumn{6}{c|}{LIVE-VQC \cite{sinno2018large}} & \multicolumn{6}{c|}{KoNViD-1k \cite{hosu2017konstanz}} &  \multicolumn{6}{c}{YouTube-UGC \cite{wang2019youtube}} \\ \cmidrule(lr){2-7}\cmidrule(lr){8-13}\cmidrule(lr){14-19}
BVQA task & \multicolumn{2}{c}{Regression} &  \multicolumn{2}{c}{Binary Class.} &  \multicolumn{2}{c|}{Ordinal Class.} & \multicolumn{2}{c}{Regression} &  \multicolumn{2}{c}{Binary Class.} &  \multicolumn{2}{c|}{Ordinal Class.} & \multicolumn{2}{c}{Regression} &  \multicolumn{2}{c}{Binary Class.} &  \multicolumn{2}{c}{Ordinal Class.} \\ \cmidrule(lr){2-3}\cmidrule(lr){4-5}\cmidrule(lr){6-7}\cmidrule(lr){8-9}\cmidrule(lr){10-11}\cmidrule(lr){12-13}\cmidrule(lr){14-15}\cmidrule(lr){16-17}\cmidrule(lr){18-19}
Model & SRCC$\uparrow$ & PLCC$\uparrow$ &   Acc.$\uparrow$ &  B.Acc.$\uparrow$ &  MZE$\downarrow$ & MAE$\downarrow$ & SRCC$\uparrow$ & PLCC$\uparrow$ &   Acc.$\uparrow$ &  B.Acc.$\uparrow$ &  MZE$\downarrow$ & MAE$\downarrow$ & SRCC$\uparrow$ & PLCC$\uparrow$ &   Acc.$\uparrow$ &  B.Acc.$\uparrow$ &  MZE$\downarrow$ & MAE$\downarrow$ \\
\midrule

BRISQUE 
& .577 & .617 & 77.3 & 72.4 & .392 & .432 
& .668 & .665 & 78.2 & 75.3 & .387 & .433
& .367 & .380 & 64.1 & 62.2 & .513 & .538 \\
GM-LOG  
& .588 & .624 & 77.0 & 72.0 & .399 & .442 
& .658 & .657 & 76.1 & 72.7 & .403 & .445 
& .350 & .367 & 62.5 & 61.6 & .518 & .552 \\

HIGRADE 
& .587 & .615 & 75.4 & 70.3 & .432 & .520 
& .701 & .708 & 79.3 & 76.7 & .396 & .431 
& \textbf{.741} & \textbf{.725} & \textbf{77.3} & \textbf{76.7} & .360 & .377 \\
FRIQUEE 
& .639 & .685 & 77.3 & 72.6 & .378 & .412
& .751 & .752 & 79.6 & 76.8 & .358 & .386
& \textbf{.756} & \textbf{.755} & \textbf{76.1} & \textbf{75.9} & \textbf{.326} & \textbf{.344} \\

CORNIA  
& .689 & .734 & 81.0 & 76.9 & .362 & .394
& .749 & .741 & 81.1 & 78.9 & .360 & .383
& .575 & .582 & 70.5 & 70.1 & .393 & .428 \\
HOSA 
& .685 & .745 & \textbf{81.4} & 77.4 & .359 & \textbf{.386} 
& \textbf{.769} & .743 & \textbf{81.4} & \textbf{79.3} & .360 & .387
& .600 & .603 & 72.0 & 71.6 & .409 & .440 \\

VGG-19 
& .622 & .712 & \textbf{81.7} & \textbf{77.7} & \textbf{.349} & \textbf{.385} 
& .708 & .729 & 80.3 & 78.1 & .383 & .413
& .539 & .553 & 73.3 & 72.5 & .414 & .442 \\
ResNet-50 
& .679 & \textbf{.747} & 80.8 & 76.1 & .396 & .446 
& \textbf{\underline{.791}} & \textbf{\underline{.799}} & \textbf{\underline{82.4}} & \textbf{\underline{79.9}} & \textbf{\underline{.342}} & \textbf{\underline{.364}}
& .721 & .717 & 74.7 & 74.3 & \textbf{.347} & \textbf{.364} \\ 
\midrule

VBLIINDS 
& \textbf{.696} & .717 & 80.5 & \textbf{77.6} & .379 & .421 
& .700 & .693 & 77.2 & 74.4 & .379 & .412
& .536 & .531 & 71.3 & 70.3 & .399 & .433 \\
TLVQM
& \textbf{\underline{.793}} & \textbf{\underline{.793}} & \textbf{\underline{82.6}} & \textbf{\underline{79.4}} & \textbf{\underline{.307}} & \textbf{\underline{.337}} 
& \textbf{.769} & \textbf{.765} & 79.3 & 76.6 & \textbf{.348} & \textbf{.376}
& .663 & .655 & 73.8 & 73.0 & .386 & .404 \\

VIDEVAL 
& \textbf{.747} & \textbf{.756} & 78.9 & 75.0 & \textbf{.354} & .397 
& \textbf{.785} & \textbf{.779} & \textbf{81.2} & \textbf{78.5} & \textbf{.350} & \textbf{.370}
& \textbf{\underline{.771}} & \textbf{\underline{.767}} & \textbf{\underline{80.0}} & \textbf{\underline{80.2}} & \textbf{\underline{.307}} & \textbf{\underline{.320}} \\
 \bottomrule
\end{tabular}
\end{table*}

\section{Experiments}
\label{sec:experiments}

\subsection{Experimental Setup}
\label{ssec:datasets}

We selected the six UGC picture and video quality datasets summarized in Table \ref{table:1} for the benchmarking experiments. Among these, CLIVE \cite{ghadiyaram2015massive}, KonIQ-10K \cite{hosu2020koniq}, and SPAQ \cite{fang2020perceptual} include authentically distorted pictures, whether manually captured or sampled from the web; and LIVE-VQC \cite{sinno2018large}, KoNViD-1k \cite{hosu2017konstanz}, and YouTube-UGC \cite{wang2019youtube} are three large scale video databases containing realistic distortions. All the datasets provide ground truth MOS as prediction targets, which we use for the quality regression task (Task A). The other two tasks, binary classification (Task B) and three-class ordinal classification (Task C) use the GMM-discretized scores as labels, based on the original MOS, as discussed in Sections \ref{task2} and \ref{task3}. Table \ref{table:1} also shows the GMM-learned thresholds used for tasks B \& C on each database, respectively. The tasks are summarized as follows.
\vspace{-1.5mm}
\begin{itemize}
\setlength\itemsep{-0.em}
  \item \textbf{Task A - Score Regression}: Given a UGC picture/video, predict a numeric quality score.
  \item \textbf{Task B - Binary Classification}: Given a UGC picture/video, predict whether it is of $\mathtt{high}$ or $\mathtt{low}$ quality.
  \item \textbf{Task C - Ordinal Classification}: Given a UGC picture/video, estimate the quality category on a three-point scale: $\{ \mathtt{Low},\mathtt{Medium},\mathtt{High}\}$.
\end{itemize}
\vspace{-1.5mm}

A set of representative BIQA/BVQA models were selected as performance references to be compared with, which include: BRISQUE \cite{mittal2012no}, GM-LOG \cite{xue2014blind}, HIGRADE \cite{kundu2017no}, FRIQUEE \cite{ghadiyaram2017perceptual}, CORNIA \cite{ye2012unsupervised}, and HOSA \cite{xu2016blind}, for both IQA and VQA datasets, and V-BLIINDS \cite{saad2014blind}, TLVQM \cite{korhonen2019two}, and VIDEVAL \cite{tu2020ugc} for VQA evaluations only. When evaluated on videos, the BIQA models were computed at one frame per second and the features average pooled across sampled frames to obtain video-level features to be used for training. We did not include any deep learning-based methods, since the model head has to be modified on our proposed tasks. However, we did utilize VGG-19 and ResNet-50 average-pooled feature maps as additional ConvNet baselines.

The performance metrics used are SRCC and PLCC for the regression (Task A), and the accuracy and balanced accuracy for the binary classification (Task B), and the mean absolute error (MAE) and mean zero-one error (MZE) for the ordinal classification (Task C). Following prior studies, we randomly divided the dataset into $80\%/20\%$ (stratified splits for classification tasks) content-disjoint training and test sets 20 times, and report the average performance on the test set. A support vector machine (SVM) \cite{chang2011libsvm} with randomized grid search cross validation was used for all tasks for a fair comparison, although one more advanced learning toolset \cite{gutierrez2015ordinal} could be contemplated for ordinal classification. For practical reasons we used a LinearSVM for CORNIA and HOSA. All the feature extractions were conducted in MATLAB while the training and evaluations were implemented with Python.

\subsection{Performance and Discussion}

Tables \ref{table:iqa} and \ref{table:vqa} show the performances of the evaluated UGC-QA models on IQA and VQA datasets for the three proposed evaluation tasks, respectively. It may be seen from Table \ref{table:iqa} that different tasks yield different rankings of the models on CLIVE - the best model is ResNet-50 for regression, FRIQUEE for binary classificion, and HOSA for ordinal classification. Similarly, FRIQUEE and ResNet-50 are the top performers for the three tasks on SPAQ. On KonIQ-10k, however, the top performing model was FRIQUEE for each task. The top three models were different for each individual task, however. This suggests that the two new tasks
provide different and complementary criteria relative to the regression task for the selection and ranking of UGC-QA models.

On the video datasets shown in Table \ref{table:vqa}, we observed more consistent results among the evaluated BVQA models - the best algorithm was TLVQM on LIVE-VQC, ResNet-50 on KoNViD-1k, and VIDEVAL on YouTube-UGC, respectively, for all the three tasks. Since any VQA dataset is small enough to contain some bias \cite{tu2020ugc, torralba2011unbiased}, there may exist models that outperform all those tested, on all tasks. But the overall performance ranking of the three different tasks still yield different results on each video set, yielding much more information than only using regression metrics.  Overall, the proposed evaluation tasks provide a different way to predict quality compared to the regression objective, with practical advantages, helping to advance studies of UGC-QA algorithms.  

\section{Conclusion}
\label{sec:conc}

We revisited the problem of no-reference quality assessment of user-generated content (UGC-QA), and proposed two additional tasks beyond the original regression approach - binary, and ordinal classification - to evaluate UGC-QA models at coarser levels. Our experimental results present reliable benchmarks on several popular UGC picture and video datasets, paving the way for further studies of UGC-QA models. We hope this work sheds insights into new views, experiments, and evaluation methods on the trending and challenging UGC-QA problem.

\vfill\pagebreak

\bibliographystyle{IEEEtran}
\footnotesize\bibliography{refs}

\begin{thebibliography}{10}
\providecommand{\url}[1]{#1}
\csname url@samestyle\endcsname
\providecommand{\newblock}{\relax}
\providecommand{\bibinfo}[2]{#2}
\providecommand{\BIBentrySTDinterwordspacing}{\spaceskip=0pt\relax}
\providecommand{\BIBentryALTinterwordstretchfactor}{4}
\providecommand{\BIBentryALTinterwordspacing}{\spaceskip=\fontdimen2\font plus
\BIBentryALTinterwordstretchfactor\fontdimen3\font minus
  \fontdimen4\font\relax}
\providecommand{\BIBforeignlanguage}[2]{{%
\expandafter\ifx\csname l@#1\endcsname\relax
\typeout{** WARNING: IEEEtran.bst: No hyphenation pattern has been}%
\typeout{** loaded for the language `#1'. Using the pattern for}%
\typeout{** the default language instead.}%
\else
\language=\csname l@#1\endcsname
\fi
#2}}
\providecommand{\BIBdecl}{\relax}
\BIBdecl

\bibitem{wang2004image}
Z.~Wang, A.~C. Bovik, H.~R. Sheikh, and E.~P. Simoncelli, ``Image quality
  assessment: from error visibility to structural similarity,'' \emph{IEEE
  transactions on image processing}, vol.~13, no.~4, pp. 600--612, 2004.

\bibitem{li2018vmaf}
Z.~Li, C.~Bampis, J.~Novak, A.~Aaron, K.~Swanson, A.~Moorthy, and J.~Cock,
  ``{VMAF}: The journey continues,'' \emph{Netflix Techn. Blog}, 2018.

\bibitem{tu2020ugc}
Z.~Tu, Y.~Wang, N.~Birkbeck, B.~Adsumilli, and A.~C. Bovik, ``{UGC-VQA}:
  Benchmarking blind video quality assessment for user generated content,''
  \emph{arXiv preprint arXiv:2005.14354}, 2020.

\bibitem{mittal2012no}
A.~Mittal, A.~K. Moorthy, and A.~C. Bovik, ``No-reference image quality
  assessment in the spatial domain,'' \emph{IEEE Trans. Image Process.},
  vol.~21, no.~12, pp. 4695--4708, 2012.

\bibitem{ghadiyaram2017perceptual}
D.~Ghadiyaram and A.~C. Bovik, ``Perceptual quality prediction on authentically
  distorted images using a bag of features approach,'' \emph{J. Vision},
  vol.~17, no.~1, pp. 32--32, 2017.

\bibitem{saad2014blind}
M.~A. Saad, A.~C. Bovik, and C.~Charrier, ``Blind prediction of natural video
  quality,'' \emph{IEEE Trans. Image Process.}, vol.~23, no.~3, pp. 1352--1365,
  2014.

\bibitem{korhonen2019two}
J.~Korhonen, ``Two-level approach for no-reference consumer video quality
  assessment,'' \emph{IEEE Trans. Image Process.}, vol.~28, no.~12, pp.
  5923--5938, 2019.

\bibitem{li2019quality}
D.~Li, T.~Jiang, and M.~Jiang, ``Quality assessment of in-the-wild videos,'' in
  \emph{Proc. ACM Multimedia Conf. (MM)}, 2019, pp. 2351--2359.

\bibitem{ebenezer2020no}
J.~P. Ebenezer, Z.~Shang, Y.~Wu, H.~Wei, and A.~C. Bovik, ``No-reference video
  quality assessment using space-time chips,'' \emph{arXiv preprint
  arXiv:2008.00031}, 2020.

\bibitem{xue2014blind}
W.~Xue, X.~Mou, L.~Zhang, A.~C. Bovik, and X.~Feng, ``Blind image quality
  assessment using joint statistics of gradient magnitude and laplacian
  features,'' \emph{IEEE Trans. Image Process.}, vol.~23, no.~11, pp.
  4850--4862, 2014.

\bibitem{ye2012unsupervised}
P.~Ye, J.~Kumar, L.~Kang, and D.~Doermann, ``Unsupervised feature learning
  framework for no-reference image quality assessment,'' in \emph{Proc. IEEE
  Conf. Comput. Vis. Pattern Recognit. (CVPR)}, 2012, pp. 1098--1105.

\bibitem{tu2020bband}
Z.~Tu, J.~Lin, Y.~Wang, B.~Adsumilli, and A.~C. Bovik, ``Bband index: a
  no-reference banding artifact predictor,'' in \emph{Proc. IEEE Int. Conf.
  Acoust., Speech, Signal Process. (ICASSP)}, 2020, pp. 2712--2716.

\bibitem{seshadrinathan2011temporal}
K.~Seshadrinathan and A.~C. Bovik, ``Temporal hysteresis model of time varying
  subjective video quality,'' in \emph{Proc. IEEE Int. Conf. Acoust., Speech,
  Signal Process. (ICASSP)}, 2011, pp. 1153--1156.

\bibitem{tu2020comparative}
Z.~{Tu}, C.~J. {Chen}, L.~H. {Chen}, N.~{Birkbeck}, B.~{Adsumilli}, and A.~C.
  {Bovik}, ``A comparative evaluation of temporal pooling methods for blind
  video quality assessment,'' in \emph{Proc. IEEE Int. Conf. Image Process.
  (ICIP)}, 2020, pp. 141--145.

\bibitem{chen2020perceptual}
L.-H. Chen, C.~G. Bampis, Z.~Li, J.~Sole, and A.~C. Bovik, ``Perceptual video
  quality prediction emphasizing chroma distortions,'' \emph{arXiv preprint
  arXiv:2009.11203}, 2020.

\bibitem{ma2017end}
K.~Ma, W.~Liu, K.~Zhang, Z.~Duanmu, Z.~Wang, and W.~Zuo, ``End-to-end blind
  image quality assessment using deep neural networks,'' \emph{IEEE Trans.
  Image Process.}, vol.~27, no.~3, pp. 1202--1213, 2017.

\bibitem{kang2014convolutional}
L.~Kang, P.~Ye, Y.~Li, and D.~Doermann, ``Convolutional neural networks for
  no-reference image quality assessment,'' in \emph{Proc. IEEE Conf. Comput.
  Vis. Pattern Recognit. (CVPR)}, 2014, pp. 1733--1740.

\bibitem{Bosse2018}
S.~Bosse, D.~Maniry, K.-R. Muller, T.~Wiegand, and W.~Samek, ``Deep neural
  networks for no-reference and full-reference image quality assessment,''
  \emph{IEEE Trans. Image Process.}, vol.~27, no.~1, pp. 206--219, Jan. 2018.

\bibitem{sheikh2006statistical}
H.~R. Sheikh, M.~F. Sabir, and A.~C. Bovik, ``A statistical evaluation of
  recent full reference image quality assessment algorithms,'' \emph{IEEE
  Trans. Image Process.}, vol.~15, no.~11, pp. 3440--3451, 2006.

\bibitem{deng2017image}
Y.~Deng, C.~C. Loy, and X.~Tang, ``Image aesthetic assessment: An experimental
  survey,'' \emph{IEEE Signal Process. Mag.}, vol.~34, no.~4, pp. 80--106,
  2017.

\bibitem{Murray2012}
N.~Murray, L.~Marchesotti, and F.~Perronnin, ``{AVA}: A large-scale database
  for aesthetic visual analysis,'' in \emph{Proc. IEEE Conf. Comput. Vis.
  Pattern Recognit. (CVPR)}, Jun. 2012.

\bibitem{zeng2019unified}
H.~Zeng, Z.~Cao, L.~Zhang, and A.~C. Bovik, ``A unified probabilistic
  formulation of image aesthetic assessment,'' \emph{IEEE Trans. Image
  Process.}, vol.~29, pp. 1548--1561, 2019.

\bibitem{ying2020patches}
Z.~Ying, H.~Niu, P.~Gupta, D.~Mahajan, D.~Ghadiyaram, and A.~Bovik, ``From
  patches to pictures (paq-2-piq): Mapping the perceptual space of picture
  quality,'' in \emph{Proc. IEEE Conf. Comput. Vis. Pattern Recognit. (CVPR)},
  2020, pp. 3575--3585.

\bibitem{lin2015experimental}
J.~Y. Lin, L.~Jin, S.~Hu, I.~Katsavounidis, Z.~Li, A.~Aaron, and C.-C.~J. Kuo,
  ``Experimental design and analysis of jnd test on coded image/video,'' in
  \emph{Appl. Digital Image Process. {XXXVIII}}, vol. 9599, 2015, p. 95990Z.

\bibitem{huynh2010study}
Q.~Huynh-Thu, M.-N. Garcia, F.~Speranza, P.~Corriveau, and A.~Raake, ``Study of
  rating scales for subjective quality assessment of high-definition video,''
  \emph{IEEE Trans. Broadcast.}, vol.~57, no.~1, pp. 1--14, 2010.

\bibitem{Wang2020}
Y.~Wang, H.~Talebi, F.~Yang, J.~G. Yim, N.~Birkbeck, B.~Adsumilli, and
  P.~Milanfar, ``Video transcoding optimization based on input perceptual
  quality,'' in \emph{Appl. Digital Image Process. {XLIII}}, A.~G. Tescher and
  T.~Ebrahimi, Eds., Aug. 2020.

\bibitem{hosu2017konstanz}
V.~Hosu, F.~Hahn, M.~Jenadeleh, H.~Lin, H.~Men, T.~Szir{\'a}nyi, S.~Li, and
  D.~Saupe, ``The konstanz natural video database (konvid-1k),'' in \emph{Proc.
  9th Int. Conf. Qual. Multimedia Exper. (QoMEX)}, 2017, pp. 1--6.

\bibitem{chang2011libsvm}
C.-C. Chang and C.-J. Lin, ``Libsvm: A library for support vector machines,''
  \emph{ACM Trans. Intell. Syst. Technol.}, vol.~2, no.~3, pp. 1--27, 2011.

\bibitem{pei2015image}
S.-C. Pei and L.-H. Chen, ``Image quality assessment using human visual dog
  model fused with random forest,'' \emph{IEEE Trans. Image Process.}, vol.~24,
  no.~11, pp. 3282--3292, 2015.

\bibitem{chen2019proxiqa}
L.-H. Chen, C.~G. Bampis, Z.~Li, A.~Norkin, and A.~C. Bovik, ``Proxiqa: A proxy
  approach to perceptual optimization of learned image compression,''
  \emph{arXiv preprint arXiv:1910.08845}, 2019.

\bibitem{ghadiyaram2015massive}
D.~Ghadiyaram and A.~C. Bovik, ``Massive online crowdsourced study of
  subjective and objective picture quality,'' \emph{IEEE Trans. Image
  Process.}, vol.~25, no.~1, pp. 372--387, 2015.

\bibitem{hosu2020koniq}
V.~Hosu, H.~Lin, T.~Sziranyi, and D.~Saupe, ``Koniq-10k: An ecologically valid
  database for deep learning of blind image quality assessment,'' \emph{IEEE
  Trans. Image Process.}, vol.~29, pp. 4041--4056, 2020.

\bibitem{yu2019predicting}
X.~Yu, C.~G. Bampis, P.~Gupta, and A.~C. Bovik, ``Predicting the quality of
  images compressed after distortion in two steps,'' \emph{IEEE Trans. Image
  Process.}, vol.~28, no.~12, pp. 5757--5770, 2019.

\bibitem{yu2020predicting}
X.~Yu, N.~Birkbeck, Y.~Wang, C.~G. Bampis, B.~Adsumilli, and A.~C. Bovik,
  ``Predicting the quality of compressed videos with pre-existing
  distortions,'' \emph{arXiv preprint arXiv:2004.02943}, 2020.

\bibitem{sinno2018large}
Z.~Sinno and A.~C. Bovik, ``Large-scale study of perceptual video quality,''
  \emph{IEEE Trans. Image Process.}, vol.~28, no.~2, pp. 612--627, 2018.

\bibitem{wang2019youtube}
Y.~Wang, S.~Inguva, and B.~Adsumilli, ``Youtube ugc dataset for video
  compression research,'' in \emph{Proc. IEEE Int. Workshop Multimedia Signal
  Process. (MMSP)}, 2019, pp. 1--5.

\bibitem{fang2020perceptual}
Y.~Fang, H.~Zhu, Y.~Zeng, K.~Ma, and Z.~Wang, ``Perceptual quality assessment
  of smartphone photography,'' in \emph{Proc. IEEE Conf. Comput. Vis. Pattern
  Recognit. (CVPR)}, 2020, pp. 3677--3686.

\bibitem{gutierrez2015ordinal}
P.~A. Guti{\'e}rrez, M.~Perez-Ortiz, J.~Sanchez-Monedero, F.~Fernandez-Navarro,
  and C.~Hervas-Martinez, ``Ordinal regression methods: survey and experimental
  study,'' \emph{IEEE Trans. Knowl. Data Eng.}, vol.~28, no.~1, pp. 127--146,
  2015.

\bibitem{kundu2017no}
D.~Kundu, D.~Ghadiyaram, A.~C. Bovik, and B.~L. Evans, ``No-reference quality
  assessment of tone-mapped {HDR} pictures,'' \emph{IEEE Trans. Image
  Process.}, vol.~26, no.~6, pp. 2957--2971, 2017.

\bibitem{xu2016blind}
J.~Xu, P.~Ye, Q.~Li, H.~Du, Y.~Liu, and D.~Doermann, ``Blind image quality
  assessment based on high order statistics aggregation,'' \emph{IEEE Trans.
  Image Process.}, vol.~25, no.~9, pp. 4444--4457, 2016.

\bibitem{torralba2011unbiased}
A.~Torralba and A.~A. Efros, ``Unbiased look at dataset bias,'' in \emph{Proc.
  IEEE Conf. Comput. Vis. Pattern Recognit. (CVPR)}, 2011, pp. 1521--1528.

\end{thebibliography}

\end{document}